%% file: root.tex
\documentclass[letterpaper, 10 pt, conference]{ieeeconf}  

\usepackage{graphics} 
\usepackage{graphicx}
\usepackage{url}
\usepackage{amsmath}
\usepackage{amsfonts}
\usepackage{xcolor}
\usepackage{listings}
\usepackage{subcaption}
\usepackage{multirow}
\usepackage{siunitx}
\usepackage{placeins}
\usepackage{algorithm}
\usepackage{algorithmicx}
\usepackage{algpseudocode}
\usepackage{amssymb}
\usepackage{bm}
\usepackage{bbm}
\usepackage{color}
\usepackage[top=1.5in,bottom=1in,right=1in,left=1in]{geometry}
\usepackage[pdfborder={0 0 0}]{hyperref}
\usepackage{empheq}
\usepackage{ulem} 

\include{macros}

\colorlet{RED}{red}
\IEEEoverridecommandlockouts                              

\title{\LARGE \bf
Zero-Shot Policy Transferability for the Control of a Scale Autonomous Vehicle
}

\author{Harry Zhang$^{1}$, Stefan Caldararu$^{2}$, Sriram Ashokkumar$^{3}$, Ishaan Mahajan$^{2}$,
\\ Aaron Young$^{4}$, Alexis Ruiz$^{1}$, Huzaifa Unjhawala$^{1}$, Luning Bakke$^{5}$ and Dan Negrut$^{6}$
\thanks{$^{1}$PhD student with the Department of Mechanical Engineering,
        University of Wisconsin - Madison, 1513 University Ave, Madison, WI, USA
        {\tt\small hzhang699@wisc.edu}}%
\thanks{$^{2}$Undergraduate Student with the Department of Computer Science,
        University of Wisconsin - Madison, 1210 W Dayton St, Madison, WI, USA
        {\tt\small scaldararu@wisc.edu}}%
\thanks{$^{3}$Masters Student with the Department of Computer Science,
        University of Wisconsin - Madison, 1210 W Dayton St, Madison, WI, USA
        }%
\thanks{$^{4}$PhD Student with the Department of Mechanical Engieering,
        Massachusetts Institute of Technology, 33 Massachusetts Ave, Cambridge, MA, USA}%
\thanks{$^{5}$Postdoc with the Department of Mechanical Engineering,
        University of Wisconsin - Madison, 1513 University Ave, Madison, WI, USA}%
\thanks{$^{6}$Professor with the Department of Mechanical Engineering,
        University of Wisconsin - Madison, 1513 University Ave, Madison, WI, USA}%
}

\begin{document}

\maketitle
\thispagestyle{empty}
\pagestyle{empty}

\begin{abstract}
We report on a study that employs an in-house developed simulation infrastructure to accomplish zero shot policy transferability for a control policy associated with a scale autonomous vehicle. We focus on implementing policies that require no real world data to be trained (Zero-Shot Transfer), and are developed in-house as opposed to being validated by previous works. We do this by implementing a Neural Network (NN) controller that is trained only on a family of circular reference trajectories. The sensors used are RTK-GPS and IMU, the latter for providing heading. The NN controller is trained using either a human driver (via human in the loop simulation), or a Model Predictive Control (MPC) strategy. We demonstrate these two approaches in conjunction with two operation scenarios: the vehicle follows a waypoint-defined trajectory at constant speed; and the vehicle follows a speed profile that changes along the vehicle's waypoint-defined trajectory. The primary contribution of this work is the demonstration of Zero-Shot Transfer in conjunction with a novel feed-forward NN controller trained using a general purpose, in-house developed simulation platform. 
\end{abstract}

\section{INTRODUCTION}
\label{sec:intro}
\input{sections/intro.tex}

\section{BACKGROUND}
\label{sec:back}
\input{sections/back.tex}

\section{PRELIMINARIES}
\label{sec:plat}
\input{sections/plat.tex}

\section{METHOD}
\label{sec:method}
\input{sections/method.tex}

\section{EXPERIMENTAL RESULTS}
\label{sec:exp}
\input{sections/exp.tex}

\section{DISCUSSION}
\label{sec:analysis}
\input{sections/analysis.tex}

\section{CONCLUSIONS}
\label{sec:conclusion}
\input{sections/conclusion.tex}

\clearpage
\newpage

\bibliographystyle{IEEEtran}
\bibliography{BibFiles/refsSensors,BibFiles/refsMachineLearning,BibFiles/refsAutonomousVehicles,BibFiles/refsChronoSpecific,BibFiles/refsSBELspecific,BibFiles/refsMBS,BibFiles/refsCompSci,BibFiles/refsTerramech,BibFiles/refsFSI,BibFiles/refsRobotics,BibFiles/refsDEM}
\end{document}

%% file: macros.tex
\newcommand{\ARTvehicle}{{\texttt{SAV}}}
\newcommand{\dtARTvehicle}{{\texttt{dtSAV}}}

\newcommand{\softpackage}[1]{{{#1}}}
\newcommand{\CHRONO}{{Chrono}}
\newcommand{\CHRONOVEH}{{\softpackage{{Chrono::Vehicle}}}}
\newcommand{\CHRONOSEN}{{\softpackage{{Chrono::Sensor}}}}
\newcommand{\Synchrono}{{\softpackage{{Synchrono}}}}

\newcommand{\vect}[1]{\mathbf{#1}}

%% file: sections/intro.tex
\subsection{Motivation}
\label{sec:mot}
Simulation can be a powerful tool for designing better robots and autonomous vehicles as it can reduce design costs, accelerate the design cycle, and enable the testing of a larger pool of candidate designs in a diverse set of scenarios difficult to reproduce in reality \cite{PNASsimRobotics2021}. The idea of using simulation in robot design is hindered by the so called simulation-to-reality, or sim2real, gap~\cite{sim2realGapEssex1995}: in many cases, an algorithm that performs well in simulation may display poor performance in reality due to hard to model aspects present in the real system~\cite{EKFMPC2023}, e.g., slackness in the steering mechanism, delays in actuation, complex sensor behavior. When Machine Learning (ML) comes into play in designing control policies, ZST~\cite{ZeroShotTransfer2020} is even more elusive since one has to make do exclusively with synthetic data. Past contributions that focused on ZST either came short of demonstrating performance in the real world~\cite{end2endMUBO2022}, or required a mixture of simulated and real data, e.g.~\cite{simBasedRL2020, VISTASimMITRL}. 


\subsection{Contribution}
\label{sec:contrib}
We outline a general purpose approach that uses our Autonomy Research Testbed (ART) platform~\cite{artatk2022} to synthesize an ML-based controller in simulation and subsequently demonstrate it on a scale vehicle, thus achieving ZST. Our contributions are threefold. \textbf{(\textit{i})} We propose the use of a physics-based high fidelity simulator to synthesize control policies without resorting to domain randomization or domain adaptation to achieve ZST. \textbf{(\textit{ii})} We do not resort to off-the-shelf proven algorithms. Rather, we train a NN via imitation learning on a low diversity dataset obtained by driving the vehicle along a small family of circles of different radii. The model is trained to imitate either a human driver or a Model Predictive Control (MPC) algorithm. Producing training data in simulation and training the NN model takes minutes. \textbf{(\textit{iii})} We demonstrate that our simulator, called Chrono~\cite{chronoOverview2016,projectChronoWebSite}, used in conjunction with ART can support autonomy stack development in simulation with good sim2real transferability traits.

\subsection{Related Work}
\label{sec:related}
Our approach is similar to \cite{end2endNVIDIA2016}, except that perception is carried out differently (GPS and IMU as opposed to camera), and we draw exclusively on simulated data.
In~\cite{end2endMUBO2022}, the authors train an all-terrain-vehicle in simulation to drive off-road while avoiding obstacles but their policy is only demonstrated in simulation. In~\cite{testFailGAN2018}, the authors focus on automatic test generation for AVs and propose ``robustness values'' for their Autonomy Stack (A-Stack), but do not provide evidence for the translation of this robustness value into real world results. AutoVRL, a platform similar to the ART one discussed herein, is presented in~\cite{autoVRL2023}, but its authors do not demonstrate real-world operation of their A-Stack. Unlike these contributions that only demonstrate performance in simulation, a body of literature shows performance in reality by focusing on combining simulation and real-world data. Most of these relate to autonomous agents that use camera and LiDAR sensors for end-to-end solutions in which an ML model directly takes sensor inputs while it outputs control commands. In~\cite{testFailGAN2018, NAMO2022} the perception training is done purely on real world datasets, i.e., KITTI~\cite{KITTI2012} and YCB~\cite{YCB2015}, respectively. Many works focus on \textit{Domain Translation} (DT) where a separate ML module is tasked with changing the appearance of sensor outputs. For instance, in~\cite{geneSISRT2017} the authors use a Generative Adversarial Network (GAN) to learn a mapping function from simulated images to a potential real world counterpart, highlighting their use of unlabeled reality data. The A-Stack is then trained on this enhanced simulation data. Similar to this, the VISTA simulator~\cite{VISTASimMITRL} uses a data-driven approach to generate sensor data, integrating the process described above into the simulator. In~\cite{VRGoggles2019}, the authors do the inverse of the operation described above. They train their A-Stack in simulation, and then generate ''VR-Goggles'' for their vehicle. When driving in reality, the sensor data is first processed by the VR-Goggles to look more like simulated data, and then passed to the A-Stack. This also requires reality data for training the goggles. While all of these approaches achieve simulation based A-Stack training with transfer to reality, they still require preprocessing of real world data for their testing, stopping short of ZST.

\textit{Domain Randomization} (DR) has been widely used to impart robustness to a control policy synthesized in simulation. This is commonly done for robotic-arm control policies, see, for instance,~\cite{ArmDomainRandomization2018}. In \cite{ZeroShotTransfer2020}, the authors discuss the effectiveness of different DR techniques. In our work, we choose to not employ DR for ZST and instead emphasize the role that an accurate simulator and model can play in accomplishing ZST. Agasint this backdrop, our effort is motivated by two observations. Firstly, the DR technique often lacks clarity on why it succeeds or fails. Secondly, DR can serve as a reliable option for enhancing robustness provided one already achieved ZST with a good simulator and model.

Finally, in this study, we showcase the integration of the Chrono simulation engine and the ART platform, both of which are being collaboratively developed with input from this group. Looking beyond Chrono, commonly used simulators include CARLA~\cite{carlaAVsim2017}, Isaac Sim~\cite{ISAACSim2018}, MuJoCo~\cite{todorovMujoco2012}, webots \cite{webots2004}, Coppelia Robotics \cite{coppeliaSim2023}, Gazebo~\cite{koenig2004design} and PyBullet~\cite{bulletPhysicsEngine2020}. Several of these solutions highlight photorealism and fast computation times, producing results that are plausible but not necessarily physically meaningful since they draw on game engines~\cite{DeterminismOfGameEngines2022}. This revokes some of the benefits of simulation based training, as one is no longer guaranteed to be able to exactly replicate physical scenarios/tests, and may make determining failure causes difficult. Finally, in addition to embracing a physics-based approach to simulation, one aspect that sets apart Chrono is its ability to embed humans in the loop and either allow them to guide the data generation process for training (as done in this contribution), or synthesize and test autonomy solutions that come into play in human-robot interaction applications.

%% file: sections/back.tex
The multi-physics simulator used used in this study is called \CHRONO~\cite{chronoOverview2016}. Two modules, \CHRONOVEH~\cite{chronoVehicle2019} and \CHRONOSEN~\cite{asherSensors2020}, provide high-fidelity, physics-based vehicle and sensor simulation, respectively, and can be leveraged for synthesizing control policies. While this work focuses on a scaled on-road car equipped with GPS and IMU sensors, \CHRONO~allows for the combination of various wheeled and tracked vehicles, and proprioceptive and exteroceptive sensor types, e.g., camera, LiDAR~\cite{artatk2022,end2endMUBO2022}. It has good terramechanics support for off-road mobility fidelity, the user having the ability to choose several terrain models~\cite{chronoSCM2019,weiTracCtrl2022}. Furthermore, the platform supports human-in-the-loop simulation, and the interaction of multiple autonomous agents, whether in intricate traffic scenarios or convoy operations utilizing \Synchrono~\cite{synchrono2018,synchrono2020}.
ART/ATK provides a ROS 2 framework that leverages the \CHRONO~simulation engine to enable autonomy algorithm synthesis in simulation followed by demonstration in the real world. A more detailed description is provided in Sec.~\ref{sec:plat_artatk}



%% file: sections/plat.tex
\subsection{ART/ATK and Chrono}
\label{sec:plat_artatk}

The control synthesis takes place in simulation in line with ZST expectations. The Chrono simulator is used to produce the time evolution of the scale vehicle.  The ART in ``ART/ATK'' provides a ROS2-based basic autonomy stack. It is implemented in Python and is Docker-containerized so that \textit{the same} autonomy stack, running on \textit{the same} processor (NVIDIA Jetson AGX card) is used both in simulation and the real world. The autonomy stack is deployed on the {\ARTvehicle} vehicle (from ``scaled Autonomous Vehicle''), which has in {\dtARTvehicle} a Chrono digital twin. Finally, the ATK component is a utility that produces the Docker container infrastructure required to accommodate the ART autonomy stack, be it in simulation or in real world testing ~\cite{artatk2022}. An IMU sensor provides heading information, and an RTK GPS delivers centimeter-scale accuracy for localization. An Extended Kalman Filter (EKF) is used for velocity estimation, utilizing the 4-DOF model described in sec.~\ref{sec:4DOF}. In simulation, \CHRONO~ provides {\dtARTvehicle} with ground truth information for the position and heading; the same EKF is used for velocity estimation.

\subsection{4-DOF VEHICLE MODEL AND ERROR STATE}
\label{sec:4DOF}
State estimation and MPC, the latter used in generating training data, call for a simple vehicle model. This model is not the Chrono {\dtARTvehicle} vehicle (which is highly nonlinear and complex), but a low fidelity replica that captures well enough {\dtARTvehicle}'s dynamics. In other words, when estimating state in simulation or producing a command via MPC, one uses a Chrono simulation, inside which we run a second simulation of the 4-DOF vehicle described in this subsection. The states of the 4-DOF vehicle model are $\vect{q} = [x, y, \theta, v]^T$,  with $x$ and $y$ representing the position in the Cartesian coordinates, respectively, heading angle $\theta$, and longitudinal velocity $v$. The commands $\mathbf{u}$ consist of a steering value $\delta$ in the range of $[-1,1]$, and a throttle input $\alpha$ in the range of $[0,1]$. The dynamics of the 4-DOF vehicle model is captured by the following differential equations:
\begin{subequations}
	\label{eq:all}
	\begin{empheq}[left={	\dot{\mathbf{q}}
			= \mathbf{f}(\mathbf{q},\mathbf{u}) =\empheqlbrace\,}]{align}
		& cos(\theta) \cdot v,
		\label{eq:1} \\
		& sin(\theta) \cdot v,
		\label{eq:2}\\
		& {v \cdot tan(\beta \delta) }/{l},
		\label{eq:3} \\ 
		& { T(\alpha, v) \cdot \gamma}  R_{w} / {I_{w}}. \label{eq:4}
	\end{empheq}
\end{subequations}	\label{equ:dynamics}
Equations (\ref{eq:1}), (\ref{eq:2}), and (\ref{eq:3}) describe a simplified bicycle model~\cite{guo2018vehicle}, where $\beta$ maps the steering command $\delta$ to the wheel steering angle, and $l$ is the wheel base of the vehicle. Assuming that the wheel has no slip, Eq.~(\ref{eq:4}) approximates the longitudinal acceleration $\dot{v}$ based on motor torque $T(\alpha, v)$, gear ratio $\gamma$, and wheel radius and moment of inertia, $R_w$ and $I_w$. See \cite{TR-2023-06} for more details.
\begin{figure}[h]
	\centering
	{{\includegraphics[width=0.3\textwidth]{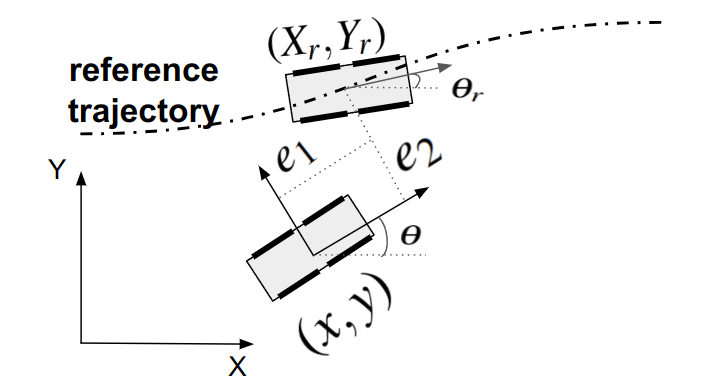} }}%
	\caption{Error state relative to target reference trajectory.}%
	\label{fig:estate}
\end{figure}
As illustrated in Fig.~\ref{fig:estate}, given a reference trajectory, the error state, $\vect{e} = [e_1, e_2, e_3, e_4]^T$, with respect to the reference one, $\vect{q}_r = [x_r, y_r, \theta_r, v_r]^T$, can be computed using Eq.~(\ref{eq:error_state})~\cite{klanvcar2007tracking}.
\begin{equation}
	\label{eq:error_state}
	\vect{e}
	=
	\begin{pmatrix}
		\cos\theta & \sin\theta & 0 & 0\\
		-\sin\theta & \cos\theta & 0 & 0\\
		0 & 0 & 1 & 0\\
		0 & 0 & 0 & 1
	\end{pmatrix}
	\begin{pmatrix}
		x_r-x\\
		y_r-y\\
		\theta_r-\theta\\
		v_r-v
	\end{pmatrix}. \;
\end{equation}
\subsection{MPC DETAILS}
\label{sec:mpc}
The MPC solution embraced is described in~\cite{TR-2023-06}. This control policy has already been tuned in simulation and tested in reality, providing insight into sim2real transferability~\cite{EKFMPC2023}. Therein, the salient conclusion was that the MPC solution was not robust, which is not an issue since it is used only to generate training data for the NN controller. The MPC is posed as
\begin{equation}
	\label{eq:error_state_discrete}
	e_{t+1} = A_t \cdot e_{t} + B_t \cdot u_t 
\end{equation}    

\begin{equation}
	\begin{aligned}
    \label{eq:opt}
    J_t^*(e_t) &= \min_{u_k} \;\; e_N^T Q e_N +\sum_{k=0}^{N-1} e_k^T Q e_k \\
	&+ (u_k-u_r)^T R (u_k-u_r) \; .
	\end{aligned}
\end{equation}


In Eq.~(\ref{eq:error_state_discrete}), we linearized the error dynamics, using~Eqs. (\ref{eq:all}) and~(\ref{eq:error_state})~\cite{TR-2023-06}. Equation~(\ref{eq:opt}) describes the optimization problem used to generate the next optimal command. For more details, please see~\cite{EKFMPC2023, TR-2023-01}.

%% file: sections/method.tex
\subsection{TRAINING DATA}
\label{sec:training}
\begin{figure}[ht]
    \centering
    {{\includegraphics[width=0.51\textwidth]{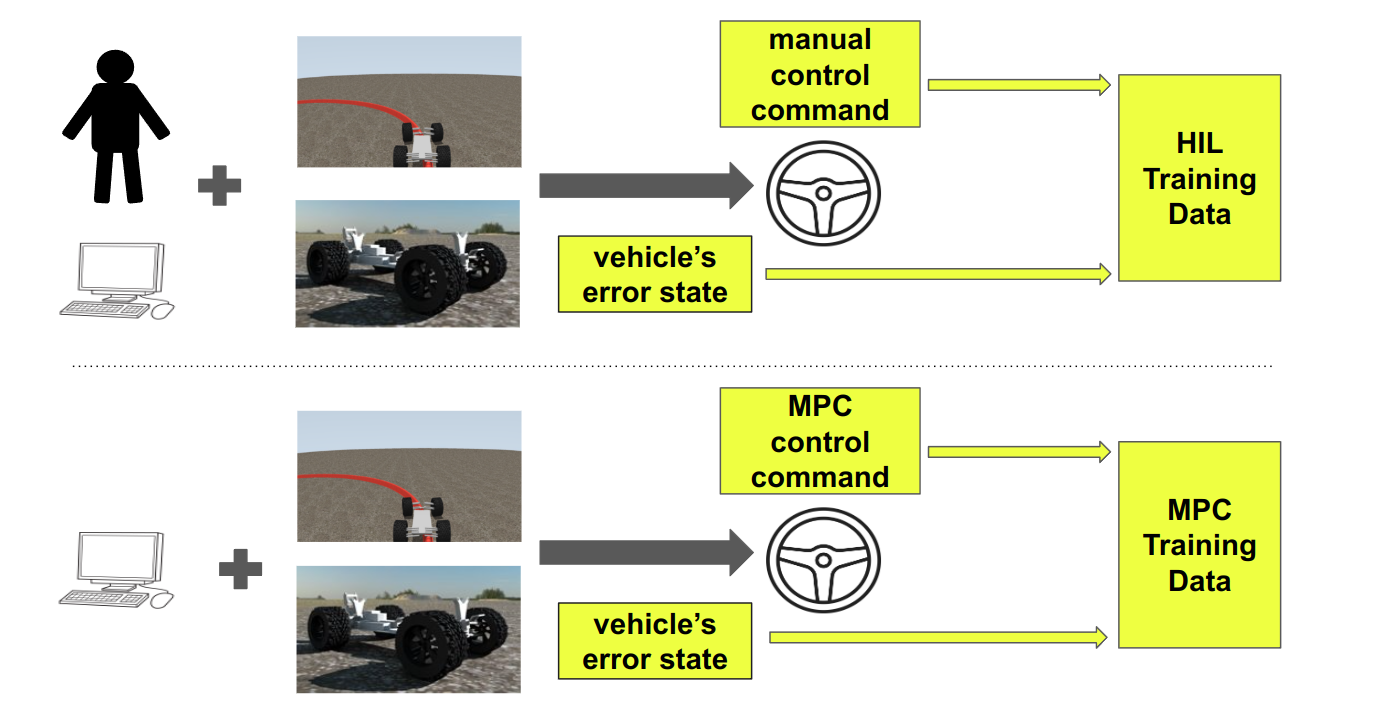} }}%
    \caption{Training Process Demonstration: upper half is the pipeline for collecting HIL (manual control) training data; lower half shows data collection using MPC.}%
    \label{fig:training}
\end{figure}

\subsubsection{Human-in-the-loop produced training data}
\label{sec:hil_training}
This approach is schematically captured in Fig.~\ref{fig:training}. Since Chrono supports human in the loop (HIL) simulation, a human drives {\dtARTvehicle} in the virtual world (manual driving) to collect data that registers what the driver does when {\dtARTvehicle} strays away from a given trajectory. We record the error state and corresponding control commands at each time step of the simulation. Seven reference trajectories are used one at a time for the training process. They are circular paths with radius 2\unit{m}, 5\unit{m}, 25\unit{m}, both clockwise and counter-clockwise, plus a 30\unit{m} straight line path, which can be thought of as a circle with infinite radius. For the multispeed control, we used seven trajectories with the same geometric shapes but different target velocities -- half of the course had a 1 \unit{m/s} velocity prescribed for the vehicle, the other half had a 2 \unit{m/s} reference velocity, with a transition velocity in between. The data helped the NN model how changes in speed elicit changes in the throttle position. Collecting training data in simulation was both simple and fast (took minutes to generate).

\subsubsection{MPC Training}
An alternative to having a human drive {\dtARTvehicle} in simulation is to use an existing MPC controller, see Sec.~\ref{sec:mpc}, and record the input commands issued by the MPC while it works to maintain a predefined trajectory. The MPC issued commands to {\dtARTvehicle} to make it follow the same reference trajectories used in the HIL data collection. The training data contains the error state and corresponding MPC command issued in each time step. The reason why \textit{our} MPC policy was not useful in reality was that it was not robust -- {\ARTvehicle} in reality did much worse than {\dtARTvehicle} in simulation, a prime manifestation of the sim2real gap.

\subsection{IMITATION LEARNING BASED CONTROLLER}

\begin{figure}
	\centering
	{{\includegraphics[width=0.5\textwidth]{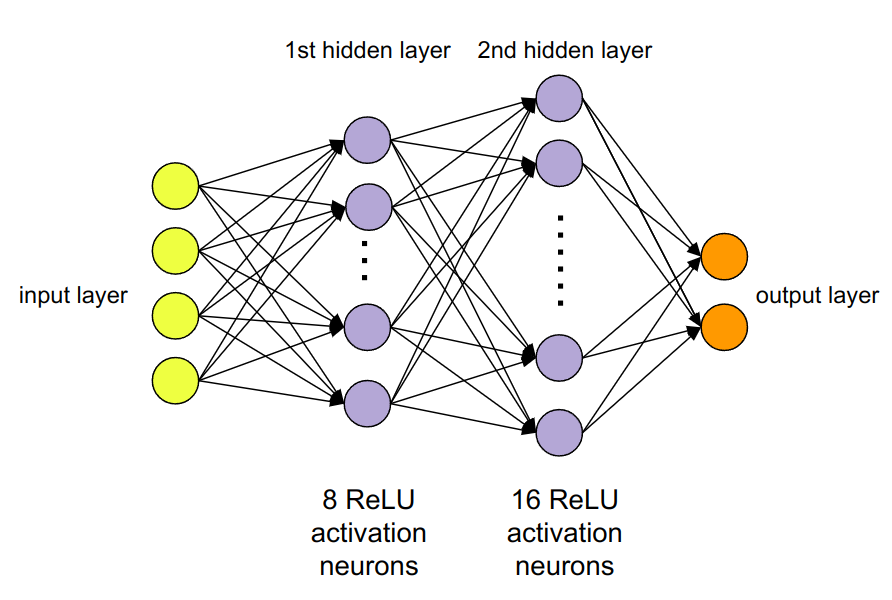} }}%
	\caption{Feed Forward Layers Setup.}%
	\label{fig:NN}
\end{figure}
We employ a feed forward Neural Network (NN) that upon training will drive {\dtARTvehicle} and subsequently {\ARTvehicle}, thus accomplishing ZST. Figure~\ref{fig:NN} depicts the two-hidden-layer NN that engages in supervised learning. Training data is generated via HIL only, or via MPC only, see Sec.~\ref{sec:training}. For NN training, the input data $\mathbf{E} \in \mathbb{R}^{4 \times n}$ is the set of error states $\mathbf{e} \in \mathbb{R}^{4 \times 1}$. This data $\mathbf{U} \in \mathbb{R}^{2 \times n}$ is matched with the control commands $\mathbf{u} \in \mathbb{R}^{2 \times 1}$. The NN is trained to produce a mapping between the error state and control command, $\mathbf{f} : \mathbf{e} \rightarrow \mathbf{u}$. The NN training and inference is carried out in Keras Core~\cite{chollet2015keras}, using PyTorch as a backend~\cite{paszke2017PyTorch}. The training converges fast since the input and output spaces are small, as are the NN's depth and width. 

%% file: sections/exp.tex
\subsection{EXPERIMENTAL SETUP}

\begin{figure}
    \centering
    {{\includegraphics[width=0.51\textwidth]{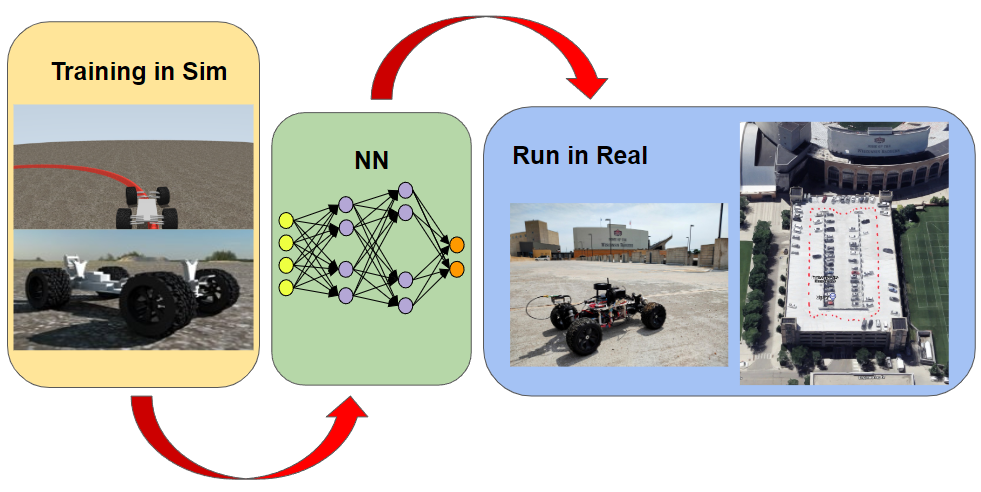} }}%
    \caption{Experiment Setup: NN was trained using {\dtARTvehicle}-generated data and subsequently tested in a parking lot using {\ARTvehicle}. The red dashed line in the right-most image represents the parking lot reference trajectory.}%
    \label{fig:real_testing}
\end{figure}

This section reports on experimental work done to assess the extent to which the control policy synthesized in simulation transferred to the real world. We established in simulation four NN policies: one trained by using a human driver, the other one using MPC-produced data; and, for each of these two scenarios we had two sub-cases: the velocity along the course was kept constant, or it changed based on the location of the {\ARTvehicle} vehicle along its trajectory. The assessment of the four NNs took place on the top of a parking lot. {\ARTvehicle} was given waypoints, and the NN controller used the IMU heading and RTK-GPS information to pass through the waypoints while following the prescribed speed regimen. The waypoints selection was mindful of the topology of the parking lot, and yielded a path that was roughly rectangular with a width by length of approximately 34 $\times$ 72~\unit[]{m}, see Fig.~\ref{fig:real_testing}. There were sinusoidal portions and regions where the speed changed along the way. For more experiment details, see uploaded movie.

\begin{figure}[h]
	\begin{subfigure}[b]{0.22\textwidth}
		\centering
		\includegraphics[width=\textwidth]{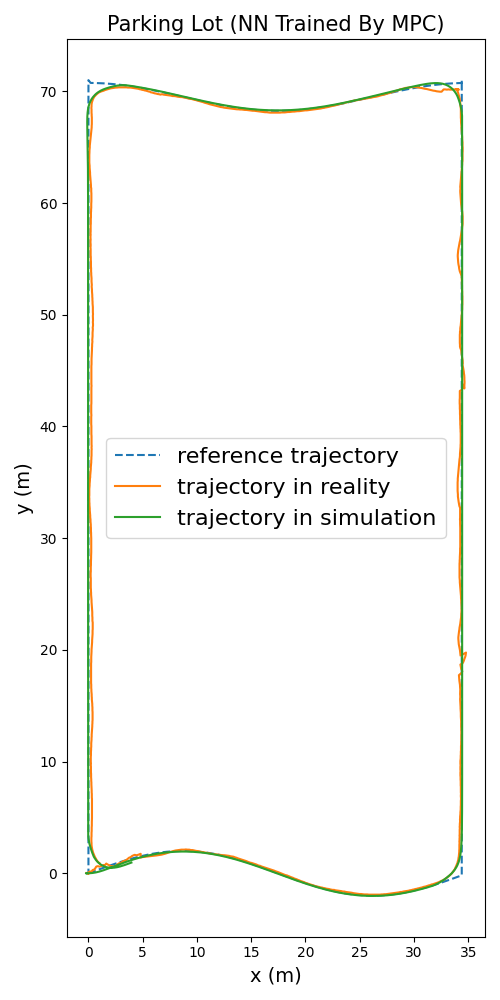} %
		\caption{}
		\label{fig:mpc_nn}
	\end{subfigure}
	\hfill
	\begin{subfigure}[b]{0.22\textwidth}
		\centering
		\includegraphics[width=\textwidth]{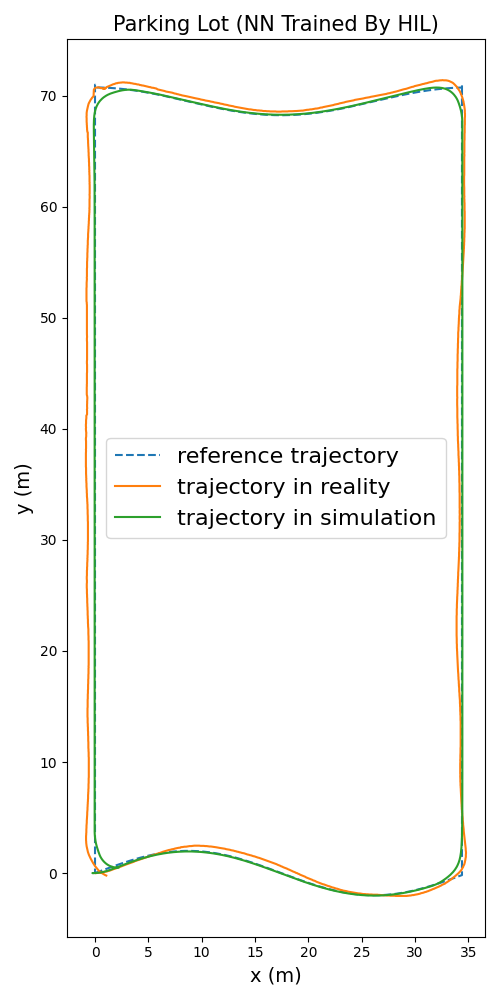} %
		\caption{}
		\label{fig:mc_nn}
	\end{subfigure}
	\hfill
	\caption{Results for simulation and reality testing: (a) NN controller trained by MPC; (b) NN controller trained by manual driving.}
	\label{fig:real_test1}
\end{figure}

\subsection{EXPERIMENTAL RESULTS} 
\subsubsection{Constant Velocity Tracking}
\begin{figure}[h]
	\begin{subfigure}[b]{0.5\textwidth}
		\centering
		\includegraphics[width=\textwidth]{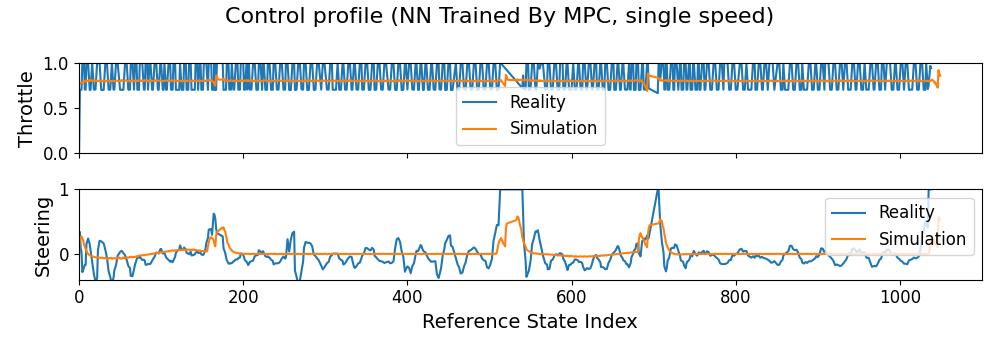} %
		\caption{}
		\label{fig:CP_NN_MPC}
	\end{subfigure}
	\hfill
	\begin{subfigure}[b]{0.5\textwidth}
		\centering
		\includegraphics[width=\textwidth]{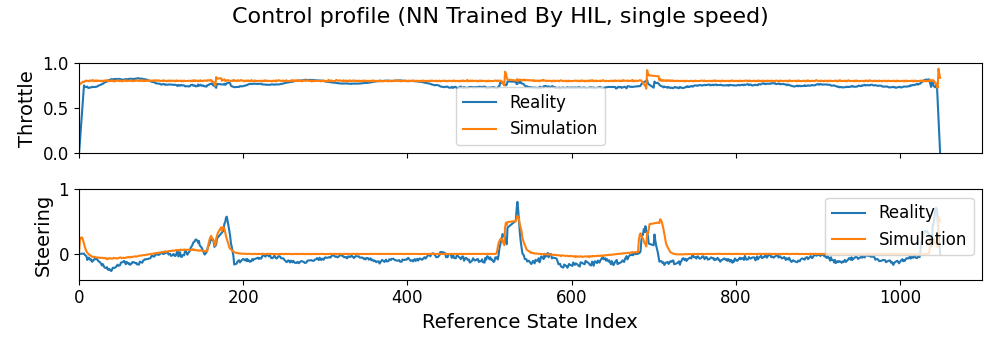} %
		\caption{}
		\label{fig:CP_NN_MC}
	\end{subfigure}
	\hfill
	\caption{Control Profiles comparing {\ARTvehicle} and {\dtARTvehicle}: (a) NN controller trained via MPC; (b) NN trained via HIL.}
	\label{fig:profiles}
\end{figure}

The reference trajectory is first followed using constant velocity along the entire path. The loop around the parking lot shown in Fig.~\ref{fig:real_testing} has a sinusoidal segment along one width, and an arc on the opposite one. Sample simulation and reality results are displayed in Fig.~\ref{fig:real_test1}. The control profiles for throttle and steering are shown in Fig.~\ref{fig:profiles}.

\begin{figure}[h]
	\begin{subfigure}[b]{0.5\textwidth}
		\centering
		\includegraphics[width=\textwidth]{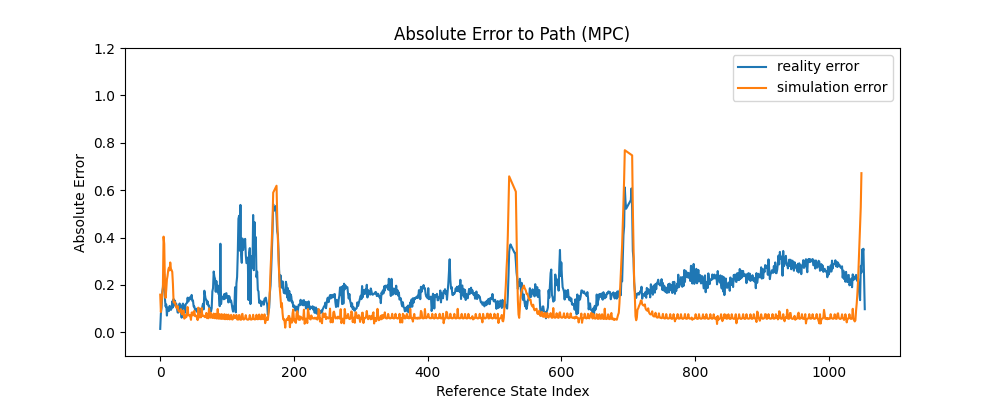} %
		\caption{}
		\label{fig:e_MPC}
	\end{subfigure}
	\hfill
	\begin{subfigure}[b]{0.5\textwidth}
		\centering
		\includegraphics[width=\textwidth]{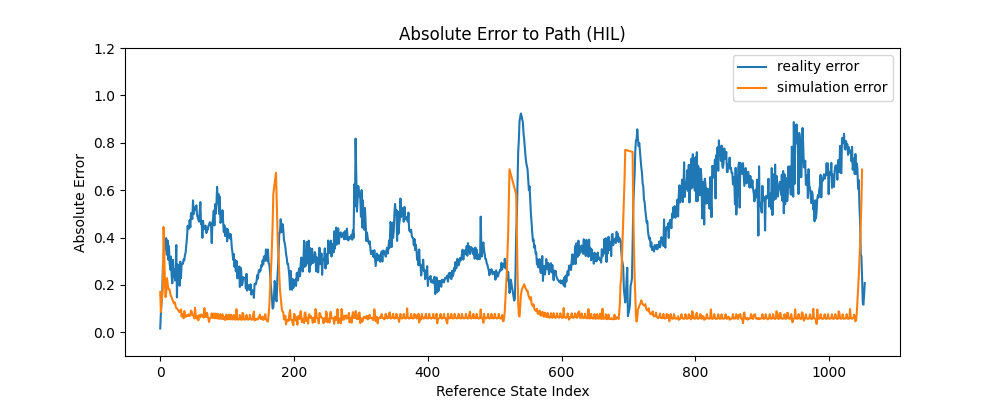} %
		\caption{}
		\label{fig:e_MC}
	\end{subfigure}
	\hfill
	\caption{Absolute sime2real errors: (a) NN controller trained by MPC; (b) NN controller trained by HIL.}
	\label{fig:errors}
	\vspace{-20pt}
\end{figure}

We ran five additional tests for each controller in each scenario. For each GPS output reading, we found the closest point on the reference trajectory, and computed the distance between the vehicle and this reference point. We then averaged these across the five tests for each reference point, and displayed them as a plot with respect to the reference location used. These absolute errors are shown in Fig.~\ref{fig:errors}. 

\subsubsection{Tracking a Complex Speed Profile}
\begin{figure}[th!]
	\begin{subfigure}[b]{0.23\textwidth}
		\centering
		\includegraphics[width=\textwidth]{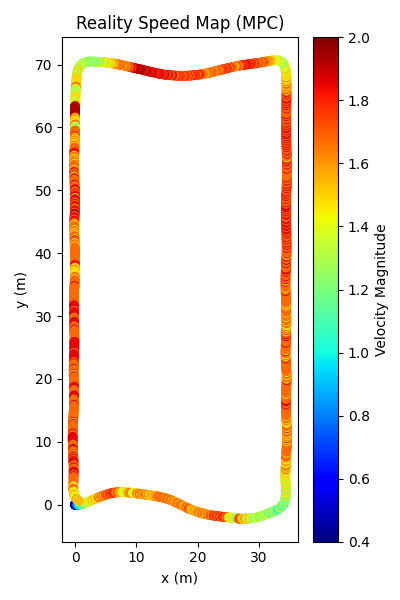} %
		\caption{}
		\label{fig:RSM_MPC}
	\end{subfigure}
	\hfill
	\begin{subfigure}[b]{0.23\textwidth}
		\centering
		\includegraphics[width=\textwidth]{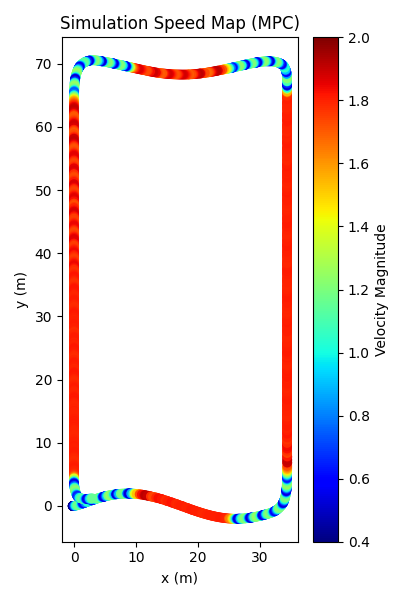} %
		\caption{}
		\label{fig:SSM_MPC}
	\end{subfigure}
	\hfill
	\begin{subfigure}[b]{0.23\textwidth}
		\centering
		\includegraphics[width=\textwidth]{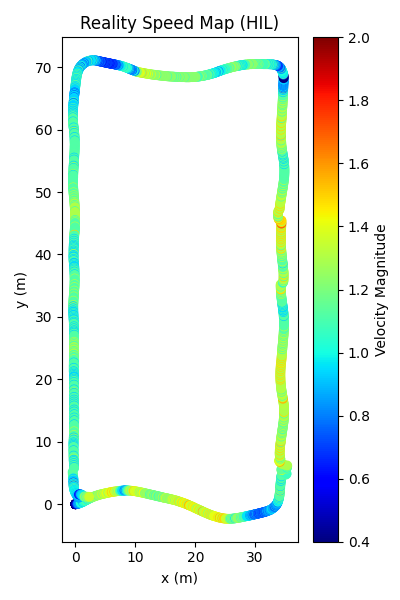} %
		\caption{}
		\label{fig:RSM_MC}
	\end{subfigure}
	\hfill
	\begin{subfigure}[b]{0.23\textwidth}
		\centering
		\includegraphics[width=\textwidth]{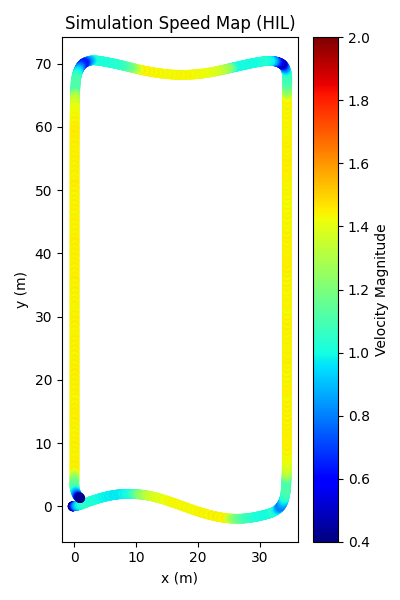} %
		\caption{}
		\label{fig:SSM_MC}
	\end{subfigure}
	\hfill
	\caption{Simulation vs. Reality Testing, variable-speed case: Fig.~\ref{fig:RSM_MPC}--{\ARTvehicle} using MPC-trained NN controller; Fig.~\ref{fig:SSM_MPC}--{\dtARTvehicle} using MPC-trained NN controller; Fig.~\ref{fig:RSM_MC}--{\ARTvehicle} using HIL-trained NN controller; Fig.~\ref{fig:SSM_MC}--{\dtARTvehicle} using HIL-trained NN controller.}
	\label{fig:multSpeed}
	\vspace{-20pt}
\end{figure}

We used the same reference trajectory, but had a speed profile that changed along the vehicle trajectory: it was 1 m/s around the four corners, and climbed to 2 m/s in between. Figure~\ref{fig:multSpeed} displays results for the two NN controllers. Heat maps show the speed of the vehicle along the trajectory, both in simulation and reality.

%% file: sections/analysis.tex
The results in Fig.~\ref{fig:real_test1} indicate that the feed-forward NN used in this study accomplishes ZST regardless of whether the synthetic training data was produced via MPC or HIL. The sim2real gap is small as demonstrated in Fig.~\ref{fig:errors}. A notable difference between simulation and reality can be noted in the commands issued, see Fig.~\ref{fig:profiles}. We hypothesize that is likely due to: a steady slant of the road for parking lot drainage purposes, which is present in reality but not in simulation; and, an unavoidable slack in the steering mechanism (with zero steering command input, the vehicle still tries to steer in the direction that the terrain is slanted towards). As shown in Fig.~\ref{fig:profiles}, for the straight line portion the reality steering has negative steering values (to compensate for the tilted road) while in simulation zero steering is maintained. Another observation from Fig.~\ref{fig:profiles} is that there is a qualitative difference between the NN trained with HIL data or MPC data. The HIL training data displays smoother changes for steering commands and prefers to merge back to the referenced trajectory slowly. Conversely, the MPC-generated training data has higher transients since it solves an optimization problem that was not instructed to account for smoothness of the ensuing maneuver. This explains why the MPC-data trained NN controller follows trajectories more precisely.

Finally, for the case when the prescribed velocity changes along the trajectory, {\dtARTvehicle} achieves a wider range of speeds compared to {\ARTvehicle}. Likewise, MPC-data training leads to a more responsive controller than when using HIL data since the MPC training takes the vehicle's powertrain model into consideration. Indeed, results in Fig.~\ref{fig:SSM_MPC} look sharper than the ones in Fig.~\ref{fig:SSM_MC}. Correspondingly, when transferring the control policies into reality, the multi-speeds control is more precise for the controller trained with MPC (in Fig.~\ref{fig:RSM_MPC}) than the one trained with HIL data (in Fig.~\ref{fig:RSM_MC}). 

Given the complexity of real-world environments, expecting exact matches between simulated and real results is exceedingly difficult to achieve. However, it is desirable to quantify this gap in safety critical applications \cite{aaronAmesSafetySim2Real2023}, and at a minimum to see traits that manifest in simulation carry over to the real world. In Figure 8, we illustrate this with multi-speed training: the HIL driver (while generating training data in the simulation) is unable to match desired speeds as precisely as the MPC. This trait is learned by the NN and appears in the real-world scenario where the NN trained on HIL doesn't match desired speeds accurately. Additionally, the HIL driver prioritizes smoother steering inputs, while the MPC prioritizes error mitigation over smoothness, as evident in the steering control inputs in Fig.~\ref{fig:profiles}.

%% file: sections/conclusion.tex
Our contributions are threefold. We demonstrated ZST for a scale vehicle using RTK-GPS and IMU sensor fusion. We established a feed-forward NN controller trained to imitate a human driver or the behavior of an MPC controller. Finally, anchored by Chrono and ART/ATK, we established an open source platform that enables the synthesis of autonomy algorithms in simulation and their demonstration in reality. The salient strength of the Chrono-ART/ATK platform is that the same ROS2 ART autonomy stack, running on the same hardware, is exercised both in simulation and reality. Since Chrono supports HIL, it enables a driver to operate a digital twin, and the data generated be subsequently used to synthesize ML-based control policies. Ongoing work focuses on increasing the determinism of the ART/ATK autonomy stack; investigating the ZST problem for a rover-like vehicle with four steerable wheels; and using simulation to synthesize autonomy stacks for ground vehicles operating on deformable terrains.

%% file: root.bbl
\def\cprime{$'$}
\begin{thebibliography}{10}
\providecommand{\url}[1]{#1}
\csname url@samestyle\endcsname
\providecommand{\newblock}{\relax}
\providecommand{\bibinfo}[2]{#2}
\providecommand{\BIBentrySTDinterwordspacing}{\spaceskip=0pt\relax}
\providecommand{\BIBentryALTinterwordstretchfactor}{4}
\providecommand{\BIBentryALTinterwordspacing}{\spaceskip=\fontdimen2\font plus
\BIBentryALTinterwordstretchfactor\fontdimen3\font minus
  \fontdimen4\font\relax}
\providecommand{\BIBforeignlanguage}[2]{{%
\expandafter\ifx\csname l@#1\endcsname\relax
\typeout{** WARNING: IEEEtran.bst: No hyphenation pattern has been}%
\typeout{** loaded for the language `#1'. Using the pattern for}%
\typeout{** the default language instead.}%
\else
\language=\csname l@#1\endcsname
\fi
#2}}
\providecommand{\BIBdecl}{\relax}
\BIBdecl

\bibitem{PNASsimRobotics2021}
\BIBentryALTinterwordspacing
H.~Choi, C.~Crump, C.~Duriez, A.~Elmquist, G.~Hager, D.~Han, F.~Hearl,
  J.~Hodgins, A.~Jain, F.~Leve, C.~Li, F.~Meier, D.~Negrut, L.~Righetti,
  A.~Rodriguez, J.~Tan, and J.~Trinkle, ``On the use of simulation in robotics:
  Opportunities, challenges, and suggestions for moving forward,''
  \emph{{Proceedings of the National Academy of Sciences}}, vol. 118, no.~1,
  2021. [Online]. Available:
  \url{https://www.pnas.org/content/118/1/e1907856118}
\BIBentrySTDinterwordspacing

\bibitem{sim2realGapEssex1995}
N.~Jakobi, P.~Husbands, and I.~Harvey, ``Noise and the reality gap: The use of
  simulation in evolutionary robotics,'' in \emph{European Conference on
  Artificial Life}.\hskip 1em plus 0.5em minus 0.4em\relax Springer, 1995, pp.
  704--720.

\bibitem{EKFMPC2023}
H.~Zhang, S.~Caldararu, I.~Mahajan, S.~Chatterjee, T.~Hansen, A.~Dashora,
  S.~Ashokkumar, L.~Fang, X.~Xu, S.~He, and D.~Negrut, ``Using simulation to
  design an mpc policy for field navigation using gps sensing,'' 2023.

\bibitem{ZeroShotTransfer2020}
E.~Valassakis, Z.~Ding, and E.~Johns, ``Crossing the gap: A deep dive into
  zero-shot sim-to-real transfer for dynamics,'' 2020.

\bibitem{end2endMUBO2022}
\BIBentryALTinterwordspacing
S.~Benatti, A.~Young, A.~Elmquist, J.~Taves, A.~Tasora, R.~Serban, and
  D.~Negrut, ``End-to-end learning for off-road terrain navigation using the
  chrono open-source simulation platform,'' \emph{Multibody System Dynamics},
  vol. \text{ }, no. https://doi.org/10.1007/s11044-022-09816-1, 2022.
  [Online]. Available: \url{https://doi.org/10.1007/s11044-022-09816-1}
\BIBentrySTDinterwordspacing

\bibitem{simBasedRL2020}
B.~Osi\'{n}ski, A.~Jakubowski, P.~Zi\c{e}cina, P.~Mi{\l}o\'{s}, C.~Galias,
  S.~Homoceanu, and H.~Michalewski, ``Simulation-based reinforcement learning
  for real-world autonomous driving,'' in \emph{2020 IEEE International
  Conference on Robotics and Automation (ICRA)}, 2020, pp. 6411--6418.

\bibitem{VISTASimMITRL}
A.~{Amini}, I.~{Gilitschenski}, J.~{Phillips}, J.~{Moseyko}, R.~{Banerjee},
  S.~{Karaman}, and D.~{Rus}, ``Learning robust control policies for end-to-end
  autonomous driving from data-driven simulation,'' \emph{IEEE Robotics and
  Automation Letters}, vol.~5, no.~2, pp. 1143--1150, 2020.

\bibitem{artatk2022}
A.~Elmquist, A.~Young, I.~Mahajan, K.~Fahey, A.~Dashora, S.~Ashokkumar,
  S.~Caldararu, V.~Freire, X.~Xu, R.~Serban, and D.~Negrut, ``A software
  toolkit and hardware platform for investigating and comparing robot autonomy
  algorithms in simulation and reality,'' \emph{arXiv preprint
  arXiv:2206.06537}, 2022.

\bibitem{chronoOverview2016}
A.~Tasora, R.~Serban, H.~Mazhar, A.~Pazouki, D.~Melanz, J.~Fleischmann,
  M.~Taylor, H.~Sugiyama, and D.~Negrut, ``{Chrono}: An open source
  multi-physics dynamics engine,'' in \emph{High Performance Computing in
  Science and Engineering -- Lecture Notes in Computer Science}, T.~Kozubek,
  Ed.\hskip 1em plus 0.5em minus 0.4em\relax Springer International Publishing,
  2016, pp. 19--49.

\bibitem{projectChronoWebSite}
{Project Chrono}, ``{Chrono}: An open source framework for the physics-based
  simulation of dynamic systems,'' \url{http://projectchrono.org}, 2020,
  accessed: 2020-03-03.

\bibitem{end2endNVIDIA2016}
M.~Bojarski, D.~Del~Testa, D.~Dworakowski, B.~Firner, B.~Flepp, P.~Goyal, L.~D.
  Jackel, M.~Monfort, U.~Muller, J.~Zhang, X.~Zhang, J.~Zhao, and K.~Zieba,
  ``End to end learning for self-driving cars,'' \emph{arXiv preprint
  arXiv:1604.07316}, 2016.

\bibitem{testFailGAN2018}
C.~E. Tuncali, G.~Fainekos, H.~Ito, and J.~Kapinski, ``Simulation-based
  adversarial test generation for autonomous vehicles with machine learning
  components,'' in \emph{2018 IEEE Intelligent Vehicles Symposium (IV)}, 2018,
  pp. 1555--1562.

\bibitem{autoVRL2023}
S.~Sivashangaran, A.~Khairnar, and A.~Eskandarian, ``Autovrl: A high fidelity
  autonomous ground vehicle simulator for sim-to-real deep reinforcement
  learning,'' 2023.

\bibitem{NAMO2022}
K.~Ellis, H.~Zhang, D.~Stoyanov, and D.~Kanoulas, ``Navigation among movable
  obstacles with object localization using photorealistic simulation,'' in
  \emph{2022 IEEE/RSJ International Conference on Intelligent Robots and
  Systems (IROS)}, 2022, pp. 1711--1716.

\bibitem{KITTI2012}
A.~Geiger, P.~Lenz, and R.~Urtasun, ``Are we ready for autonomous driving? the
  kitti vision benchmark suite,'' in \emph{2012 IEEE Conference on Computer
  Vision and Pattern Recognition}, 2012, pp. 3354--3361.

\bibitem{YCB2015}
B.~Calli, A.~Walsman, A.~Singh, S.~Srinivasa, P.~Abbeel, and A.~M. Dollar,
  ``Benchmarking in manipulation research: Using the yale-cmu-berkeley object
  and model set,'' \emph{IEEE Robotics \& Automation Magazine}, vol.~22, no.~3,
  pp. 36--52, 2015.

\bibitem{geneSISRT2017}
G.~J. Stein and N.~Roy, ``Genesis-rt: Generating synthetic images for training
  secondary real-world tasks,'' 2017.

\bibitem{VRGoggles2019}
J.~Zhang, L.~Tai, P.~Yun, Y.~Xiong, M.~Liu, J.~Boedecker, and W.~Burgard,
  ``Vr-goggles for robots: Real-to-sim domain adaptation for visual control,''
  2019.

\bibitem{ArmDomainRandomization2018}
X.~B. Peng, M.~Andrychowicz, W.~Zaremba, and P.~Abbeel, ``Sim-to-real transfer
  of robotic control with dynamics randomization,'' in \emph{2018 IEEE
  International Conference on Robotics and Automation (ICRA)}, 2018, pp.
  3803--3810.

\bibitem{carlaAVsim2017}
A.~Dosovitskiy, G.~Ros, F.~Codevilla, A.~Lopez, and V.~Koltun, ``{CARLA}: {An}
  open urban driving simulator,'' in \emph{Proceedings of the 1st Annual
  Conference on Robot Learning}, 2017, pp. 1--16.

\bibitem{ISAACSim2018}
J.~Liang, V.~Makoviychuk, A.~Handa, N.~Chentanez, M.~Macklin, and D.~Fox,
  ``Gpu-accelerated robotic simulation for distributed reinforcement
  learning,'' 2018.

\bibitem{todorovMujoco2012}
E.~Todorov, T.~Erez, and Y.~Tassa, ``{MuJoCo}: A physics engine for model-based
  control,'' in \emph{2012 IEEE/RSJ International Conference on Intelligent
  Robots and Systems}.\hskip 1em plus 0.5em minus 0.4em\relax IEEE, 2012, pp.
  5026--5033.

\bibitem{webots2004}
O.~Michel, ``Cyberbotics ltd. webots: professional mobile robot simulation,''
  \emph{International Journal of Advanced Robotic Systems}, vol.~1, no.~1,
  p.~5, 2004.

\bibitem{coppeliaSim2023}
\relax {Coppelis Robotics}, ``\url{http://www.coppeliarobotics.com/},'' 2023.

\bibitem{koenig2004design}
N.~P. Koenig and A.~Howard, ``Design and use paradigms for {Gazebo}, an
  open-source multi-robot simulator.'' in \emph{2004 IEEE/RSJ International
  Conference on Intelligent Robots and Systems (IROS)}, vol.~4.\hskip 1em plus
  0.5em minus 0.4em\relax Citeseer, 2004, pp. 2149--2154.

\bibitem{bulletPhysicsEngine2020}
E.~Coumans and Y.~Bai, ``{PyBullet}, a {Python} module for physics simulation
  for games, robotics and machine learning,'' \url{http://pybullet.org},
  2016--2019.

\bibitem{DeterminismOfGameEngines2022}
G.~Chance, A.~Ghobrial, K.~McAreavey, S.~Lemaignan, T.~Pipe, and K.~Eder, ``On
  determinism of game engines used for simulation-based autonomous vehicle
  verification,'' \emph{IEEE Transactions on Intelligent Transportation
  Systems}, vol.~23, no.~11, pp. 20\,538--20\,552, 2022.

\bibitem{chronoVehicle2019}
R.~Serban, M.~Taylor, D.~Negrut, and A.~Tasora, ``{Chrono::Vehicle}
  template-based ground vehicle modeling and simulation,'' \emph{Intl. J. Veh.
  Performance}, vol.~5, no.~1, pp. 18--39, 2019.

\bibitem{asherSensors2020}
A.~Elmquist and D.~Negrut, ``Methods and models for simulating autonomous
  vehicle sensors,'' \emph{IEEE Transactions on Intelligent Vehicles}, vol.~5,
  pp. 684--692, 2020.

\bibitem{chronoSCM2019}
A.~Tasora, D.~Mangoni, D.~Negrut, R.~Serban, and P.~Jayakumar, ``Deformable
  soil with adaptive level of detail for tracked and wheeled vehicles,''
  \emph{International Journal of Vehicle Performance}, vol.~5, no.~1, pp.
  60--76, 2019.

\bibitem{weiTracCtrl2022}
W.~Hu, Z.~Zhou, S.~Chandler, D.~Apostolopoulos, K.~Kamrin, R.~Serban, and
  D.~Negrut, ``Traction control design for off-road mobility using an
  {SPH}-{DAE} co-simulation framework,'' \emph{Multibody System Dynamics},
  vol.~55, pp. 165--188, 2022.

\bibitem{synchrono2018}
D.~Negrut, R.~Serban, A.~Elmquist, and D.~Hatch, ``Synchrono: An open-source
  framework for physics-based simulation of collaborating robots,'' in
  \emph{2018 IEEE International Conference on Simulation, Modeling, and
  Programming for Autonomous Robots (SIMPAR)}.\hskip 1em plus 0.5em minus
  0.4em\relax IEEE, 2018, pp. 101--107.

\bibitem{synchrono2020}
J.~Taves, A.~Elmquist, A.~Young, R.~Serban, and D.~Negrut, ``Synchrono: A
  scalable, physics-based simulation platform for testing groups of autonomous
  vehicles and/or robots,'' in \emph{2020 IEEE/RSJ International Conference on
  Intelligent Robots and Systems (IROS)}.\hskip 1em plus 0.5em minus
  0.4em\relax IEEE, 2020, pp. 2251--2256.

\bibitem{guo2018vehicle}
H.~Guo, D.~Cao, H.~Chen, C.~Lv, H.~Wang, and S.~Yang, ``Vehicle dynamic state
  estimation: State of the art schemes and perspectives,'' \emph{IEEE/CAA
  Journal of Automatica Sinica}, vol.~5, no.~2, pp. 418--431, 2018.

\bibitem{TR-2023-06}
H.~Zhang, H.~Unjhawala, S.~Caldararu, I.~Mahajan, L.~Bakke, R.~Serban, and
  D.~Negrut, ``Simplified 4dof bicycle model for robotics applications,''
  Simulation-Based Engineering Laboratory, University of Wisconsin-Madison,
  Tech. Rep., 2023,
  \url{https://sbel.wisc.edu/wp-content/uploads/sites/569/2023/06/TR-2023-06.pdf}.

\bibitem{klanvcar2007tracking}
G.~Klan{\v{c}}ar and I.~{\v{S}}krjanc, ``Tracking-error model-based predictive
  control for mobile robots in real time,'' \emph{Robotics and autonomous
  systems}, vol.~55, no.~6, pp. 460--469, 2007.

\bibitem{TR-2023-01}
H.~Zhang, S.~Chatterjee, T.~Hansen, S.~Caldararu, I.~Mahajan, N.~Batagoda,
  L.~Fang, R.~Serban, and D.~Negrut, ``Formulating model predictive control
  (mpc) strategies in conjunction with error dynamics based waypoint-seeking to
  model robust vehicle control,'' Simulation-Based Engineering Laboratory,
  University of Wisconsin-Madison, Tech. Rep., 2023,
  \url{https://sbel.wisc.edu/wp-content/uploads/sites/569/2023/03/TR-2023-01.pdf}.

\bibitem{chollet2015keras}
F.~Chollet \emph{et~al.}, ``Keras,'' \url{https://keras.io}, 2015.

\bibitem{paszke2017PyTorch}
A.~Paszke, S.~Gross, S.~Chintala, G.~Chanan, E.~Yang, Z.~DeVito, Z.~Lin,
  A.~Desmaison, L.~Antiga, and A.~Lerer, ``Automatic differentiation in
  {PyTorch},'' in \emph{NIPS 2017 Workshop Autodiff}, 2017, Conference
  Proceedings.

\bibitem{aaronAmesSafetySim2Real2023}
P.~Akella, W.~Ubellacker, and A.~D. Ames, ``Safety-critical controller
  verification via sim2real gap quantification,'' in \emph{2023 IEEE
  International Conference on Robotics and Automation (ICRA)}.\hskip 1em plus
  0.5em minus 0.4em\relax IEEE, 2023, pp. 10\,539--10\,545.

\end{thebibliography}
